\documentclass[a4paper,twocolumn]{article}
\pdfoutput=1

\usepackage{cite}
\usepackage[margin=1.25in]{geometry}
\usepackage{url}            
\usepackage{booktabs}       
\usepackage{amsfonts}       
\usepackage{nicefrac}       
\usepackage{microtype}      
\usepackage{dblfloatfix}    
\usepackage{graphicx}
\usepackage[subrefformat=parens,labelformat=parens]{subcaption}
\usepackage{authblk}
\usepackage{amsmath,amsthm,amssymb}
\usepackage{mathtools}
\mathtoolsset{showonlyrefs}
\usepackage{placeins}
\usepackage{enumitem}
\usepackage[font=small,labelfont=bf]{caption}

\title{Application of generative autoencoder in \textit{de novo} 
molecular design}
\date{November 20, 2017}

\author[1,2,$\dagger$]{Thomas Blaschke}
\author[1]{Marcus Olivecrona}
\author[1]{Ola Engkvist}
\author[2]{J\"urgen Bajorath}
\author[1,$\dagger$]{Hongming Chen}
\affil[1]{Hit Discovery, Discovery Sciences, Innovative Medicines and Early
Development Biotech Unit, AstraZeneca R\&D Gothenburg, 431 83 M\"olndal, Sweden}
\affil[2]{Bonn Aachen International Center for Information Technology BIT, University of Bonn, Dahlmannstrasse 2, 53113 Bonn, Germany}
\affil[$\dagger$]{thomas.blaschke@uni-bonn.de, hongming.chen@astrazeneca.com}

\begin{document}
\maketitle
\begin{abstract}
A major challenge in computational chemistry is the generation of novel molecular structures with desirable pharmacological and physiochemical properties. In this work, we investigate the potential use of autoencoder, a deep learning methodology, for \textit{de novo} molecular design. Various generative autoencoders were used to map molecule structures into a continuous latent space and vice versa and their performance as structure generator was assessed. Our results show that the latent space preserves chemical similarity principle and thus can be used for the generation of analogue structures. Furthermore, the latent space created by autoencoders were searched systematically to generate novel compounds with predicted activity against dopamine receptor type 2 and compounds similar to known active compounds not included in the trainings set were identified.
\end{abstract}

\section{Introduction}
Over the last decade, deep learning (DL) technology has been successfully applied in various areas of artificial intelligence research. DL has evolved from artificial neural networks and has often shown superior performance compared to other machine learning algorithms in areas such as image or voice recognition and natural language processing. Recently, DL has been successfully applied to different research areas in drug discovery. 
One notable application has been the use of a fully connected deep neural network (DNN) to build quantitative structure-activity relationship (QSAR) models that have outperformed some commonly used machine learning algorithms. \cite{ma_deep_2015} Another interesting application of DL is training various types of neural networks to build generative models for generating novel structures. Segler \textit{et al.} \cite{segler_generating_2017} and Yuan \textit{et al.} \cite{yuan_chemical_2017} applied a recurrent neural network (RNN) to a large number of chemical structures represented as SMILES strings to obtain generative models which learn the probability distribution of characters in a SMILES string. The resulting models were capable of generating new strings which correspond to chemically meaningful SMILES.
\begin{figure*}[ht]
    \centering
    \includegraphics[width=\linewidth, page=1]{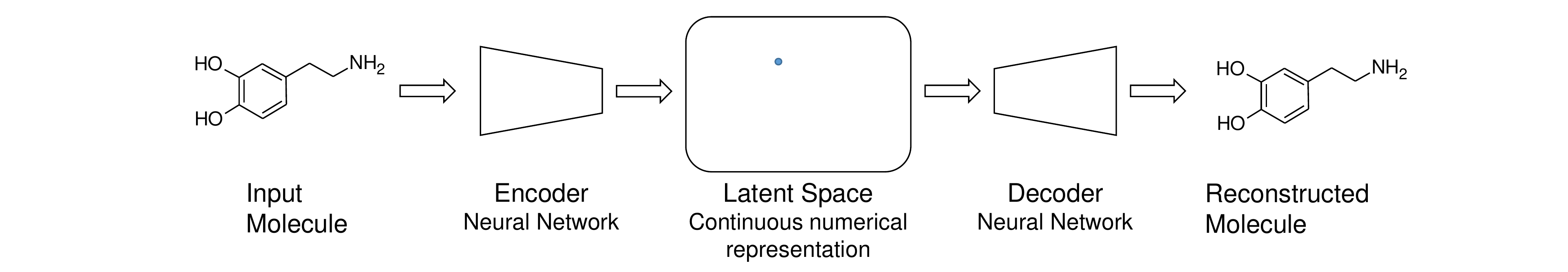}
    \caption{An autoencoder is a coordinated pair of NNs. The encoder converts a high- dimensional input, e.g. a molecule, into a continuous numerical representation with fixed dimensionality. The decoder reconstructs the input from the numerical representation.}
    \label{fig:fig1}
\end{figure*}
Jaques \textit{et al.} \cite{jaques_sequence_2016} combined RNN with a reinforcement learning method, deep Q-learning, to train models that can generate molecular structures with desirable property values for cLogP and quantitative estimate of drug-likeness (QED). \cite{bickerton_quantifying_2012} Olivecrona \textit{et al.} \cite{olivecrona_molecular_2017} proposed a policy based reinforcement learning approach to tune the pretrained RNNs for generating molecules with  user defined properties. The method has been successfully applied on inverse QSAR problems, i.e. generating new structures under the constraint of a QSAR model. Such generative DL approach allows addressing the inverse-QSAR problem more directly. The predominant inverse-QSAR approach consists of three steps. \cite{churchwell_signature_2004 , wong_constructive_2009} First, a forward QSAR model is built to fit biological activity with chemical descriptors. Second, given the forward QSAR function an inverse-mapping function is defined to map the activity to chemical descriptor values, so that a set of molecular descriptor values resulting high activity can be obtained. Third, the obtained molecular descriptor values are then translated into new compound structures. How to define an explicit inverse-mapping function transforming descriptor value to chemical structure is a major challenge for such inverse-QSAR methods. Miyao \textit{et al.} applied Gaussian mixture models to address inverse-QSAR problem, but their method can only be applied to linear regression or multiple linear regression QSAR models for the inverse-mapping. \cite{miyao_inverse_2016} This will largely limit the usage of the method, since most popular machine learning models are non-linear. \cite{douali_neural_2003} To address this issue, Miyao \textit{et al.} recently applied differential evolution to obtain optimized molecular descriptors for non-linear QSAR models, \cite{miyao_exploring_2017} though, the identification of novel compounds using virtual screening remained challenging. In contrast, the generative DL approach allows to directly generate desirable (determined by the forward QSAR models) molecules, without the need of using an explicit inverse-mapping function.
 
Autoencoder (AE) is a type of NN for unsupervised learning. It first encodes an input variable into latent variables and then decodes the latent variables to reproduce the input information. G\'ómez-Bombarelli \textit{et al.} proposes a novel method \cite{gomez-bombarelli_automatic_2016} using variational autoencoder (VAE) to generate chemical structures. After training the VAE on a large number of compound structures, the resulting latent space becomes a generative model. Sampling on the latent vectors results in chemical structures. By navigating in the latent space one could specifically search for latent points with desired chemical properties. However, a targeted search was difficult in the case study of designing organic light-emitting diodes. \cite{gomez-bombarelli_automatic_2016} Based upon G\'ómez-Bombarelli’s work, Kadurin \textit{et al.} \cite{kadurin_drugan:_2017} used VAE as a descriptor generator and coupled it with a generative adversarial network (GAN) \cite{goodfellow_generative_2014}, a special NN architecture, to identify new structures that were proposed to have desired activities.
In our current study, we have constructed various types of AE models and compared their performance in structure generation. Furthermore, Bayesian optimization was used to search for new compounds in the latent space guided by user defined target functions. This strategy was used successfully to generate new structures that were predicted to be active by a QSAR model. 

\section{Methods}
\subsection{Neural networks}
In this study we combined different NN architectures to generate molecular structures. This section provides some background information for all relevant architectures used in this work. 
\begin{figure*}[ht]
    \centering
    \includegraphics[width=\linewidth, page=2]{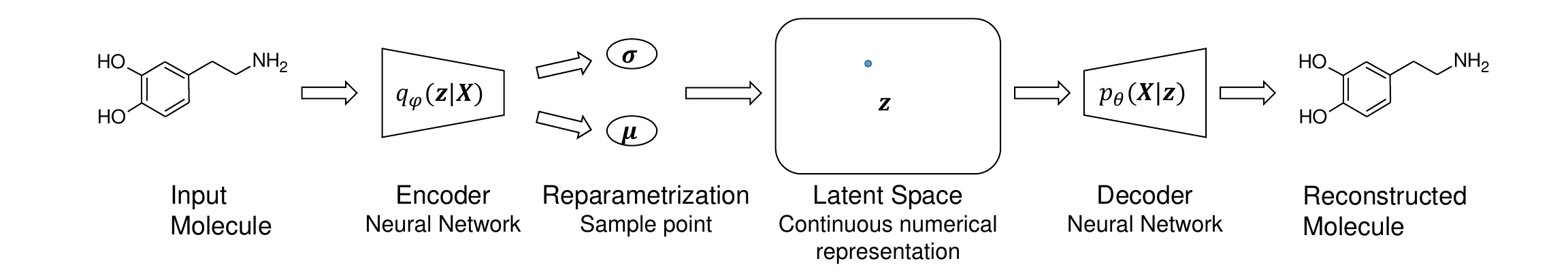}
    \caption{Encoding and decoding of a molecule using a variational autoencoder. The encoder converts a molecule structure into a Gaussian distribution deterministically. Given the generated mean and variance, a new point is sampled and fed into the decoder. The decoder then generates a new molecule from the sampled point.}
    \label{fig:fig2}
\end{figure*}
\subsubsection{Autoencoder}\label{sec:autoencoder}
Autoencoder (AE) is a NN architecture for unsupervised feature extraction. A basic AE consists of an encoder, a decoder and a distance function (Figure~\ref{fig:fig1}). The encoder is a NN that maps high-dimensional input data to a lower dimensional representation (latent space), whereas the decoder is a NN that reconstructs the original input given the lower dimensional representation. A distance function quantifies the information loss derived from the deviation between the original input and the reconstructed output. The goal of the training is to minimize the information loss of the reconstruction. Because target labels for the reconstruction are generated from the input data, the AE is regarded as self-supervised.

The input mapping to and from a lower dimensional space introduces an information bottleneck so that the AE is forced to extract informative features from the high dimensional input. The most common dimensionality reduction technique introducing an information bottleneck is principal component analysis (PCA). \cite{abdi_principal_2010} In fact, the basic AE version is sometimes referred as non-linear PCA. \cite{scholz_nonlinear_2008} The dimensionality reduction performance and the usefulness of the extracted features depend on the input data and the actual architecture of the encoder and decoder NN. It has been shown that recurrent and convolutional NNs can successfully generate text sequences and molecular structures. \cite{gomez-bombarelli_automatic_2016 , duvenaud_convolutional_2015}

\subsubsection{Recurrent Neural Networks}
Recurrent NNs (RNNs) are popular architectures with high potential for natural language processing. \cite{ian_goodfellow_deep_2016} The idea behind RNNs is to apply the same function for each element of sequential data, with the output being depended on the result of the previous step. RNN computations can be rationalized using the concept of a cell. For any given step \textit{t}, the cell \textit{t} is a result of the previous cell \textit{t−1} and the current input \textit{x}. The content of cell \textit{t} will determine both the output of the current step and influence the next cell state. This enables the network to memorize past events and model data according to previous inputs. This concept implicitly assumes that the most recent events are more important than early events since recent events influence the content of the cell the most. However, this might not be an appropriate assumption for all data sequences, therefore, Hochreiter \textit{et al.} \cite{hochreiter_long_1997} introduced the Long-Short-Term Memory cell. Through a more controlled flow of information, this cell type can decide which previous information to retain and which to discard. The Gated Recurrent Unit (GRU) is a simplified implementation of the Long-Short-Term Memory architecture and achieves much of the same effect at a reduced computational cost. \cite{chung_empirical_2014}

\subsubsection{Convolutional Neural Networks}
Convolutional NNs (CNNs) are common neural NNs for pattern recognition in images or feature extraction from text. \cite{lei_molding_2015} A CNN consists of an input and output layer as well as multiple hidden layers. The hidden layers are convolutional, pooling or fully connected. Details of different layers in CNN can be found in Simard \textit{et al.} \cite{simard_best_2003} The key feature of a CNN is introducing of convolution layers. In each layer, the same convolution operation is applied on each input data which is then replaced by a linear combination of its neighbours. The parameters of the linear combination are referred as filter or kernel, whereas the number of considered neighbours is referred as filter size or kernel size. The output of applying one filter onto the input information is called a feature map. Applying non-linearity such as sigmoid or scaled exponential linear units (SELU) \cite{klambauer_self-normalizing_2017} on a feature map allows to model nonlinear data. Furthermore, applying another CNN on top of a feature map will allow one to model features of spatially separated inputs. The pooling layer is used to reduce the size of feature maps. After passing through multiple convolution and pooling layers, the feature maps are concatenated into fully connected layers where every neuron in neighbouring layers are all connected to give final output value.
\begin{figure*}[ht]
    \centering
    \includegraphics[width=\linewidth, page=3]{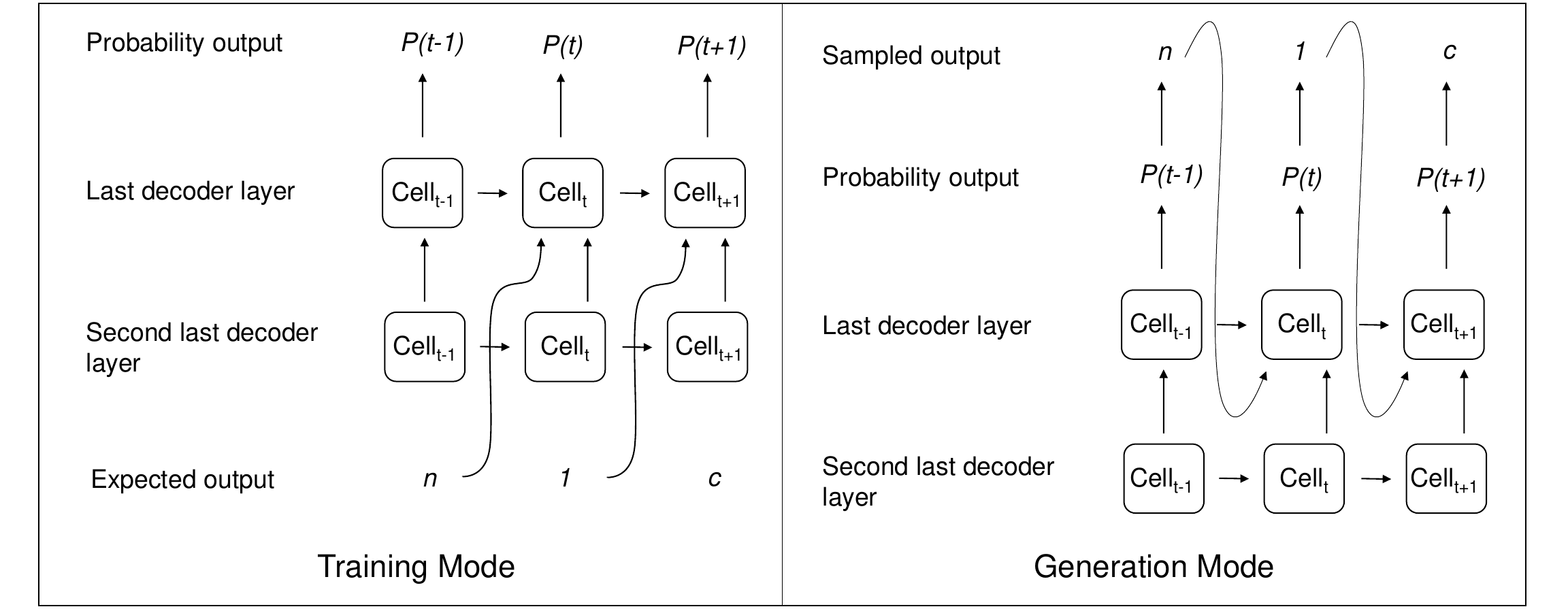}
    \caption{Sequence generation using teachers forcing. The last decoder trained with teachers forcing receives two inputs: the output of the previous layer and a character from the previous time step. In the training mode, the previous character is equal to the corresponding character from the input sequence, regardless of the probability output. During the generation mode the decoder samples at each time step a new character based on the output probability and uses this as input for the next time step.}
    \label{fig:fig3}
\end{figure*}
\subsection{Implementation details}
\subsubsection{Variational autoencoder}
The basic AE described in section~\ref{sec:autoencoder} maps a molecule $X$ into a continuous space $z$ and the decoder reconstructs the molecule from its continuous representation. However, using this basic definition, the model is not enforced to learn a generalized numeric representation of the molecules. Due to the large amount of parameters of the NNs and the comparatively small amount of training data, the AE will likely learn some explicit mapping of the training set and thus the decoder will not be able to decode arbitrary points in the continuous space. To avoid learning an explicit mapping, we restrict the model to learn a latent variable from its input data.
The variational autoencoder (VAE) \cite{kingma_auto-encoding_2013} provides a formulation in which the continuous representation $z$ is interpreted as a latent variable in a probabilistic generative model. Let $p(z)$ be the prior distribution imposed on the continuous representation, $p_{\theta}(X|z)$ be a probabilistic decoding distribution and $q_{\phi}(z|X)$ be a probabilistic encoding distribution (shown in Figure~\ref{fig:fig2}).

The parameters of $p_{\theta}(X|z)$ and $q_{\phi}(z|X)$ can be inferred during the training of the VAE via backpropagation. During training, the reconstruction error of the decoder is reduced by maximizing the log-likelihood $p_{\theta}(X|z)$. Simultaneously, the encoder is regularized to approximate the latent variable distribution $p(z)$ by minimizing the Kullback-Leibler divergence $D_{KL}(q_{\phi}(z|X) || p(z))$. If the prior $p(z)$ has to follow a multivariate Gaussian distribution with zero mean and unit variance, the loss function can be formulized as:
\begin{equation*}
L = -D_{KL}(q_{\phi}(z|X)||N(0,I)) + E[log{\ }p_{\theta}(X|z) ]     
\end{equation*}
Our VAE is implemented using the PyTorch package \cite{noauthor_pytorch_2017} and follows Gómez-Bombarelli architecture closely. \cite{gomez-bombarelli_automatic_2016} Like G\'ómez-Bombarelli \textit{et al.} we used three CNN layer followed by two fully connected neural layers as an encoder. Instead of using rectified linear units \cite{kingma_auto-encoding_2013} SELU \cite{klambauer_self-normalizing_2017} was used as non-linear activation function allowing a faster convergence of the model. The output of the last encoder layer is interpreted as mean and variance of a Gaussian distribution. A random point is sampled using these parameters and used as input for the decoder.  
The decoder employs a fully connected neural layer followed by three layers of RNNs, built using GRU cells. The last GRU layer defines the probability distribution over all possible outputs at each position in the output sequence. As shown in Figure~\ref{fig:fig3}, in the training mode, the input of the last layer is a concatenation of the output of the previous layer and token from the target SMILES (in training set), while in the generation mode, the input of the last layer is a concatenation of the output of the previous layer and the sampled token from the previous GRU cell (Figure~\ref{fig:fig3}). This method is called teachers forcing  \cite{williams_learning_1989} and is known to increase the performance on long text generation tasks.  \cite{jaeger_herbert_tutorial_2002} For comparison, we also trained a VAE model without teachers forcing, where the input of last GRU layer was not influenced by any previous token but only depended on the output of the previous GRU layer in both training and generation mode. 
\subsubsection{Adversarial autoencoder}

\begin{figure*}[ht]
    \centering
    \includegraphics[width=\linewidth, page=4]{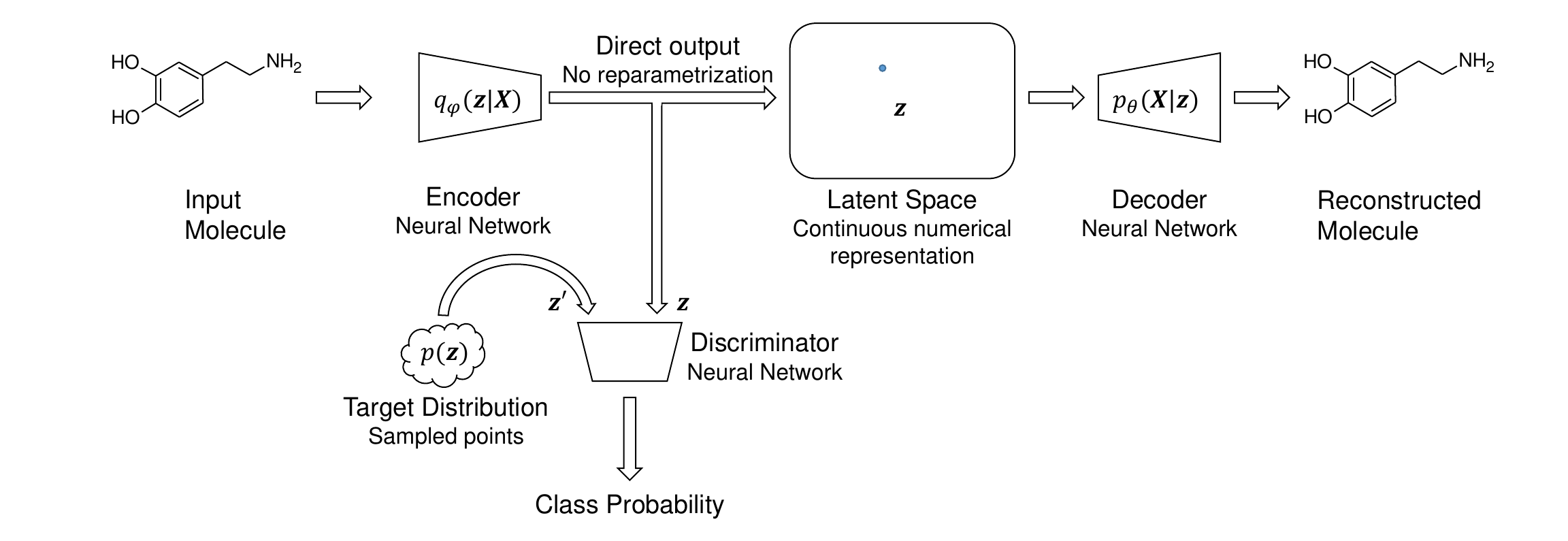}
    \caption{Learning process of an adversarial autoencoder. The encoder converts a molecule directly into a numerical representation. During training the output is not only fed into the decoder but also into a discriminator. The discriminator is trained to distinguish between the output of the encoder and a randomly sampled point from a prior distribution. The encoder is trained to “fool” the discriminator by mimicking the target prior distribution.}
    \label{fig:fig4}
\end{figure*}

The VAE assumes a simplistic prior, i.e. a Gaussian distribution, on the latent representation to keep the Kullback-Leibler divergence mathematically computable. To evaluate alternative priors we implemented a modified VAE architecture known as adversarial autoencoder (AAE).  \cite{makhzani_adversarial_2015} The major difference between the AAE and VAE is that an additional discriminator NN is added into the architecture to force the output of encoder $q_\phi (z|X)$ to follow a specific target distribution, while at the same time the reconstruction error of decoder is minimized. Figure~\ref{fig:fig4} illustrates the workflow of the AAE model. 
To regularize the encoder posterior $q_\phi (z|X)$ to match a specific prior $p(z)$, a discriminator $D$ was introduced. It consists of three fully connected NN layers. The first layer applies only an affine matrix transformation, whereas the second layer additionally applies SELU as non-linear activation. The third layer composes an affine matrix transformation followed by a sigmoid function. The discriminator $D$ receives the output $z$ of the encoder and randomly sampled points $z'$ from the target prior as input. The whole training process was done in three sequential steps:
\begin{enumerate}
\item	The encoder and decoder were trained simultaneously to minimize the reconstruction loss of the decoder:  
\begin{equation*}
    L_{p_{\theta}}= E[log{\ } p_{\theta}(X|z) ]
\end{equation*}

\item The discriminator NN D was then trained to correctly distinguish the true input signals $z'$ generated from target distribution from the false signals $z$ generated by the encoder by minimizing the loss function:
\begin{equation*}
L_{D}  = -(log(D(z'))+ log(1-D(z)))
\end{equation*}

\item In the end, the encoder NN was trained to fool the discriminator by minimizing another loss function:
\begin{equation*}
L_{q_\phi}  = -log(D(z))
\end{equation*}
\end{enumerate}
The three steps were iteratively run for each batch until the reconstruction loss function $L_{p_{\theta}}$ was converged. For all AAE models, teachers forcing scheme was always used for the decoder NN.

\subsubsection{Bayesian optimization of molecules}
To search for molecules with desired properties in latent space Bayesian optimization (BO) was implemented using the GPyOpt package.  \cite{the_gpyopt_authors_gpyopt:_2016} Here the score estimating the probability of being active against a specific target was used as the objective function and it was maximized during the BO.  
Gaussian process (GP) models  \cite{rasmussen_gaussian_2005} were trained with 100 starting points to predict the score of each molecule from the latent space. A new point was then selected by sequentially maximizing the expected improvement acquisition  \cite{jones_efficient_1998} based on the GP model. The new data point was transformed into a corresponding SMILES strings  \cite{noauthor_daylight_nodate} using the above mentioned decoder network and scored accordingly. The new latent point was added to the GP model as an additional point with associated score. The process was repeated 500 iterations to search for optimal solutions.

\subsubsection{Tokenizing SMILES and generating new SMILES}\label{subsub:token}
\begin{figure}[ht]
    \centering
    \includegraphics[width=\linewidth, page=5]{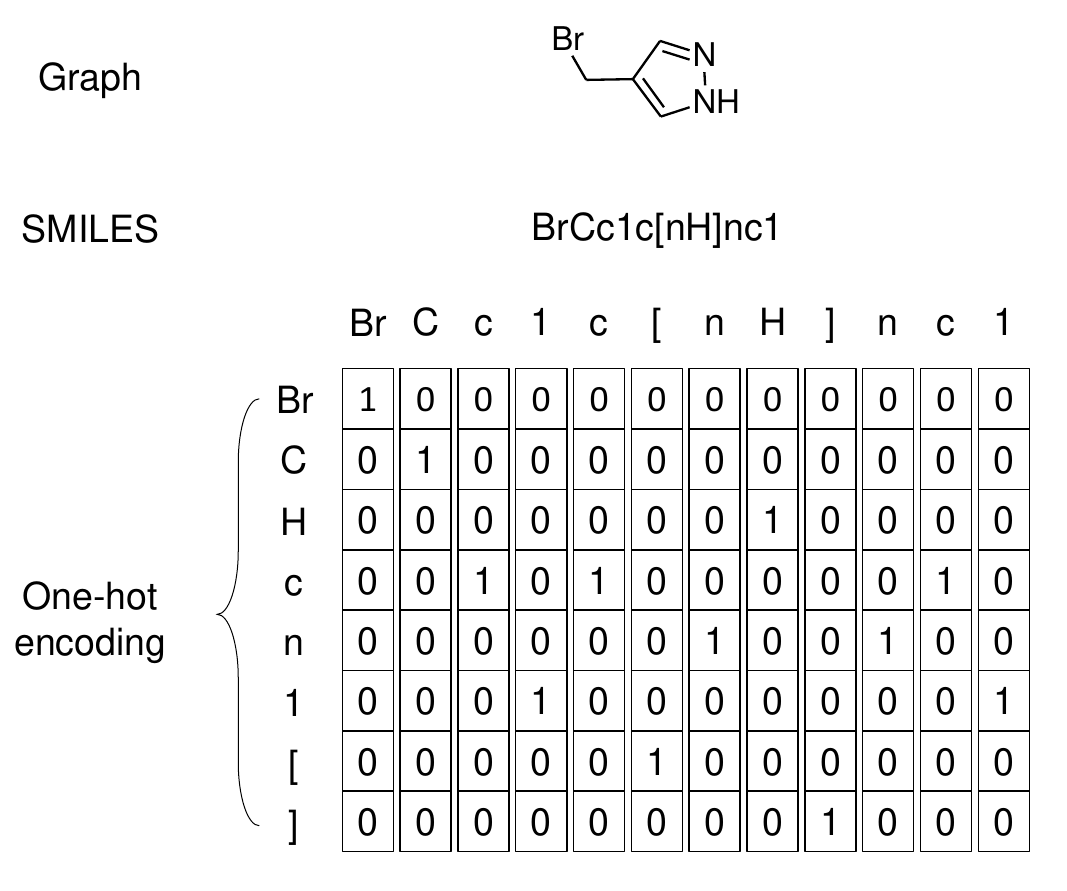}
    \caption{Different representations of 4-(bromomethyl)-1H-pyrazole. Exemplary generation of the one-hot representation derived from the SMILES. For simplicity only a reduced vocabulary is shown here, while in practice a larger vocabulary that covers all tokens present in the training data is used.}
    \label{fig:fig5}
\end{figure}

A SMILES string represents a molecule as a sequence of characters corresponding to atoms as well as special characters denoting opening and closure of rings and branches. The SMILES string is, in most cases, tokenized based on single characters, except for atom types which comprises two characters such as "Cl" and "Br", where they are considered as one token. This method of tokenization resulted in 35 tokens present in the training data. The SMILES were canonicalized using RDKit  \cite{noauthor_rdkit_2017} and encoded up to a maximum length of 120 token. Shorter SMILES were padded with spaces at the end to the same length. The sequence of token is converted into a one-hot representation and used as input for the AE. Figure~\ref{fig:fig5} illustrates the tokenization and the one-hot encoding.

Once an AE is trained, the decoder is used to generate new SMILES from arbitrary points of the latent space. Because the output of last layer of the decoder is a probability distribution over all possible tokens, the output token at each step was sampled 500 times. Thereby, we obtained 500 sequences containing 120 tokens for each latent point. The sequence of tokens was then transformed into a SMILES string and its validity was checked using RDKit. The most frequently sampled valid SMILES was assigned as the final output to the corresponding latent point. 

\subsubsection{Training of AE models}
Various AE models were trained on structures taken from ChEMBL version 22.  \cite{gaulton_chembl:_2012} The SMILES were canonicalized using RDKit and the stereochemistry information was removed for simplicity. We omitted all structures with less than 10 heavy atoms and filtered out structures that had more than 120 tokens (see Section~\ref{subsub:token}). Additionally, all compounds reported to be active against the dopamine type 2 receptor (DRD2) were removed from the set. The final set contained approx. 1.3 million unique compounds, from which we use used 1.2 million compounds as training set and the remaining 162422 compounds as a validation set. 
All AE models were trained to map to a 56-dimensional latent space. Mini-batch size of 500, a learning rate of $3.1*10^{-4}$ and stochastic gradient optimization method ADAM  \cite{kingma_adam:_2014} were used to train all models until convergence.

\subsubsection{DRD2 activity model}
A crucial objective for \textit{de novo} molecule design is to generate molecules having a high probability of being active against a given biological target. In current study, DRD2 was chosen as the target, and the same data set and activity model generated in our previous study[6] were used here. The data set was extracted from ExCAPE-DB  \cite{sun_excape-db:_2017} and contained 7218 actives ($pIC50 > 5$) and 343204 inactives ($pIC50 < 5$). A support vector machine (SVM) classification model with Gaussian kernel was built using Sci-Kit Learn \cite{noauthor_scikit-learn:_2017} on the DRD2 training set using the extended connectivity fingerprint with a diameter of 6 (ECFP6).  \cite{rogers_extended-connectivity_2010}

\section{Results and Discussion}
\begin{table*}[t]
    \caption{Reconstruction accuracy for the different AE models.}
    \centering
    \small
    \begin{tabular}{p{2.8cm}@{\hskip8pt}p{3.0cm}@{\hskip9pt}p{3.0cm}@{\hskip9pt}p{3.0cm}}
    \toprule
    \textbf{Model}   & 
    \raggedright \textbf{Average character reconstruction \% in training set (training mode)} &
    \raggedright \textbf{Average character reconstruction \% in validation set (generation mode)} & 
\raggedright \textbf{Valid SMILES \% in validation set (generation mode)}
    \tabularnewline
    \midrule
   NoTeacher VAE      & 96.8 & 96.3 & 19.3 \\
   Teacher VAE      & 97.4 & 86.2 & 77.6 \\
   Gauss AAE      & 98.2 & 89.0 & 77.4 \\
   Uniform AAE      & 98.9 & 88.5 & 78.3 \\
   \bottomrule
    \end{tabular}
    \label{tab:reconstruction}
    \vskip4pt
\end{table*}
\begin{table*}[!b]
\vskip1pt
    \caption{Exemplary sequence reconstruction.}
    \centering
    \small
    \begin{tabular}{l@{\hskip10pt}l@{\hskip10pt}l}
    \toprule
         & \textbf{Generated Sequence} & \textbf{Valid} \\
    \textbf{Target sequence} & Cc1ccc2c(c1)sc1c(=O)[nH]c3ccc(C(=O)NCCCN(C)C)cc3c12 & \\
    \midrule
    NoTeacher VAE	& Cc1ccc2cnc1)sc1c(=O)[nH]c3ccc(C(=O)NCCCN(C)C)c33c12 &	No \\
Teacher VAE	& Cc1ccc2c(c1)sc1c(=O)[nH]c3ccc(C(=O)NCCN(C)C)cc3c12 & 	Yes\\
Gauss AAE	& Cc1ccc2c(c1)sc1c(=O)[nH]c3ccc(C(=O)NCCCN(C)C)cc3c12 & Yes\\
Uniform AAE	& Cc1ccc2c(c1)sc1c(=O)[nH]c3ccc(C(=O)NCCCN(C)C)cc3c12 & Yes\\
    \bottomrule
    \end{tabular}
    \label{tab:example}
\vskip4pt
\end{table*}
In this work, we tried to address three questions: First, if compounds can be mapped into a continuous latent space and subsequently reconstructed by autoencoder NN? Second, if the latent space preserves chemical similarity principles? Third, if the latent space encoding chemical structures can be used to search and optimize structures with respect to some complex properties such as predicted biological activity?
In current study we trained and compared four different AE types: A variational autoencoder which does not use teachers forcing (named as NoTeacher VAE), a variational autoencoder which utilize teachers forcing (named as Teacher VAE) and two adversarial autoencoder where the encoder was trained to follow either a Gaussian or a Uniform distribution (named Gauss AAE and Uniform AAE).

\subsection{Structure generation}
The encoder and decoder of the AE model were trained simultaneously on the training set by minimizing the character reconstruction error of the input SMILES sequence. Once the models were trained, the validation set was first mapped into the latent space via the encoder NN and the structures were reconstructed through the decoder (i.e. generation mode). To evaluate the performance of the autoencoder, the reconstruction accuracy (percentage of position-to-position correct character) and percentage of valid SMILES string (according to RDKit SMILES definition) of whole validation set were examined.
\setcounter{figure}{6}
\begin{figure*}[!b]
  \centering
  \begin{subfigure}[b]{0.5\linewidth}
    \centering\includegraphics[width=\linewidth, page=7]{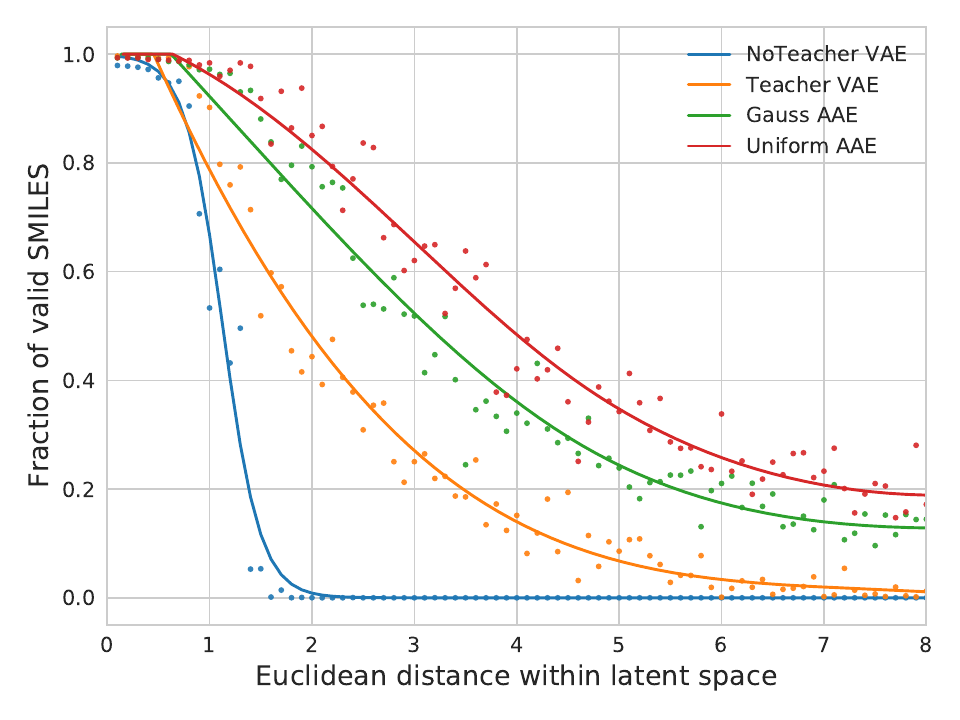}
    \caption{\label{fig:fig7a}}
  \end{subfigure}%
  \begin{subfigure}[b]{0.5\linewidth}
    \centering\includegraphics[width=\linewidth, page=8]{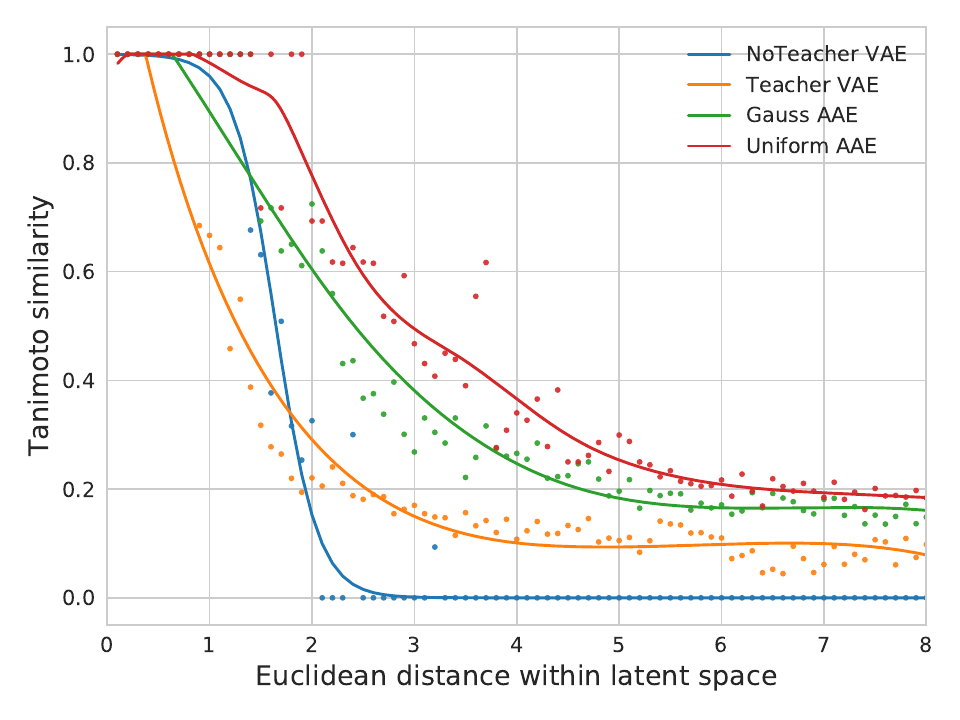}
    \caption{\label{fig:fig7b}}
  \end{subfigure}
  \caption{\subref{fig:fig7a} Chemical similarity (Tanimoto, ECFP6) of generated structures to Celecoxib in relation to the distance in the latent space. \subref{fig:fig7b} Fraction of valid SMILES generated during the reconstruction.}
  \label{fig:fig7}
\end{figure*}
The results are shown in Table~\ref{tab:reconstruction}. All methods yield good performance on character reconstruction of the SMILES sequence in the training mode and at least 95\% of all characters are correct in a reconstructed SMILES string, whereas all teachers forcing methods display a minor improved accuracy of about 1-2\% compared to NoTeacher VAE. As to the generation mode, all teachers forcing based models show decreased accuracy. This is not surprising since the information of a wrongly sampled character can propagate through the remaining sequence generation and influences the next steps. Both AAE methods achieved higher accuracy than the Teacher VAE, indicating that the decoder for these method depends more on the information from the latent space than on the previously generated character.

Interestingly teacher forcing based models demonstrate much higher percentage of valid SMILES when compared to NoTeacher VAE model, although the NoTeacher VAE method has higher character reconstruction accuracy. It means the reconstruction errors in NoTeacher VAE model are more likely to result in invalid SMILES. Browsing these invalid SMILES reveals that the NoTeacher VAE model is often not able to generate matching pairs of the branching characters "(" and ")" or ring symbols like "1" as exemplified in Table~\ref{tab:example}. Losing these ring forming and branching information makes it impossible for the NoTeacher VAE model to adapt its remaining sequence generation to remedy the errors. 

The teacher forcing containing AEs produce a significantly higher fraction of valid SMILES. They finish more often with a syntactically correct SMILES, as the information about the previously generated character influences the next generation steps. This allows the model to learn the syntax of the SMILES and thus have better chance to generate correct sequence. Their lower reconstruction accuracy in generation mode, is due to the fact that the generator must continue to generate a sequence under the condition of a previously incorrectly sampled character and thus the error propagates through the remaining sequence generation process and create additional errors. Both adversarial models show a slightly higher character reconstruction of the validation set than the Teacher VAE.

\subsection{Does the latent space preserve the similarity principle?}
To use generative AE in a \textit{de novo} molecule design task, the latent space must preserve the similarity principle, i.e. the more similar structures are, the closer they must be positioned in latent space. When the similarity principle is preserved in latent space, searching for new structures based on a query structure becomes feasible. 

A known drug, Celecoxib, was chosen as an example for validating the similarity principle in latent space. It was first mapped into latent spaces generated by various AE models and then chemical structures were sampled from the latent vectors. Figure~\ref{fig:fig6} shows the structures generated from the latent vector of Celecoxib for different AE models.
\setcounter{figure}{5}
\begin{figure}[ht]
    \centering
    \includegraphics[width=\linewidth, page=6]{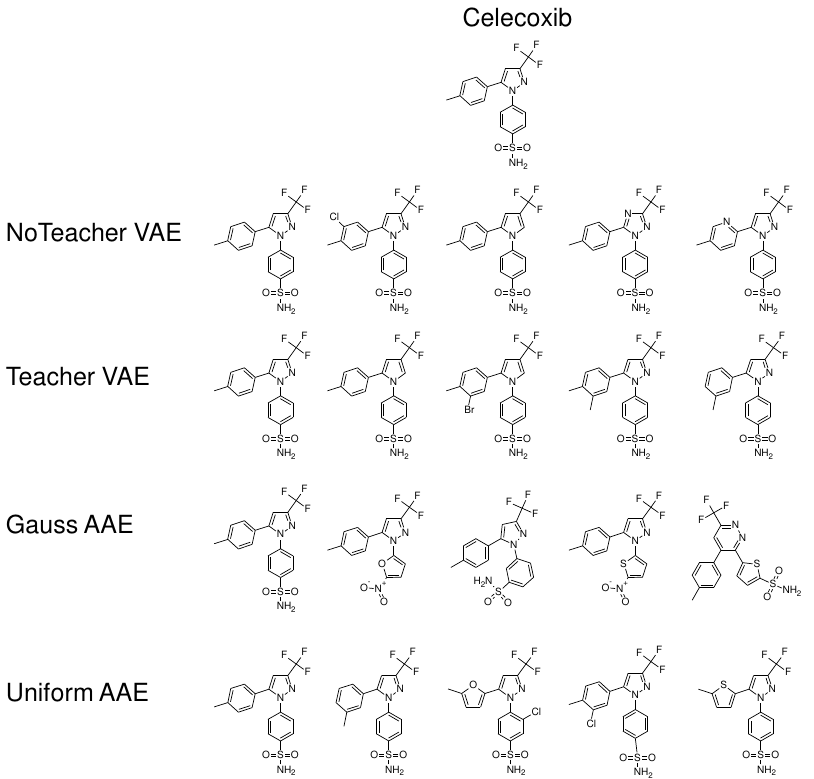}
    \caption{Sampled structures at the latent vector corresponding to Celecoxib. The structures are sorted by the relative generation frequencies in descending order from left to right.}
    \label{fig:fig6}
\end{figure}
\setcounter{figure}{7}
AE generated many analogues to Celecoxib that often only differed by a single atom.
To investigate if the similarity principle is also applicable over a larger area, we gradually increased the distance between the random latent vectors and Celecoxib from 0 to 8 with a step size of 0.1 steps. At each distance bin we sampled 10 random points and generated 500 sequences from each points, resulting in 5000 reconstruction attempts at each distance bin. The median ECFP6 Tanimoto similarity of all valid structures and Celecoxib was calculated. 

The results are shown in Figure~\ref{fig:fig7a} and demonstrate that compounds with decreasing similarity to Celecoxib generally have larger distances to Celecoxib in all four different latent spaces. This clearly indicates that the similarity principle is reserved in the surrounding area of Celecoxib making it possible to carry out similarity searches in the latent space. Figure~\ref{fig:fig7b} shows the relationship between the proportion of valid SMILES and the distance to Celecoxib in latent spaces. Uniform AAE has higher fraction of valid SMILES compared to other models. NoTeacher VAE has a very low fraction of valid SMILES at distance larger than 2. This may explain its much steeper similarity curve in Figure~\ref{fig:fig7a} compared to other models.
\begin{figure*}[ht]
  \centering
  \begin{subfigure}[b]{0.5\linewidth}
    \centering\includegraphics[width=\linewidth, page=9]{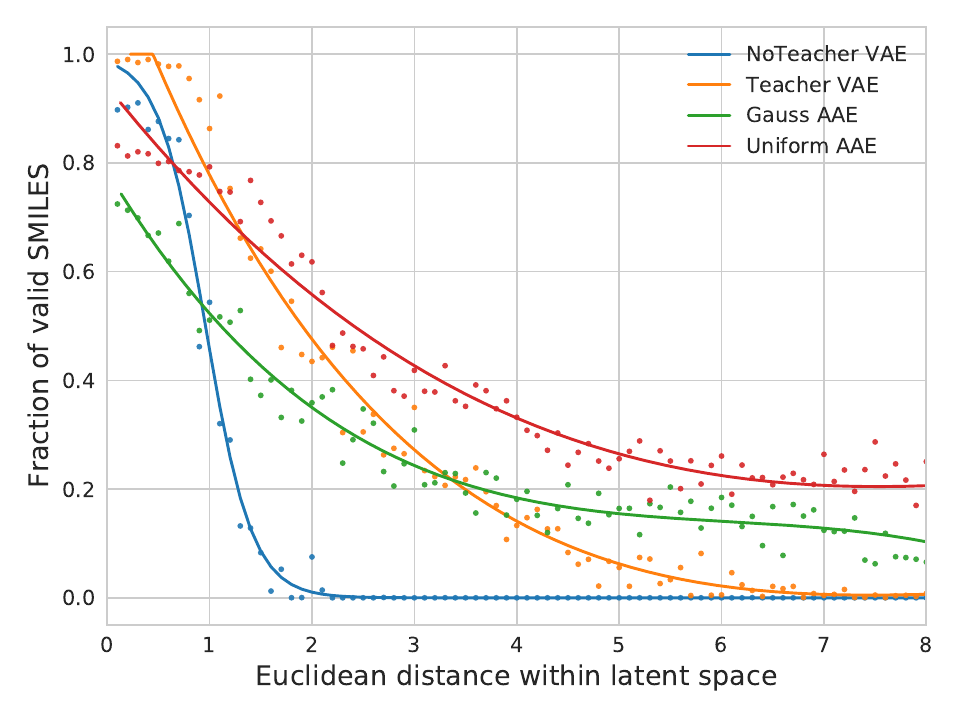}
    \caption{\label{fig:fig8a}}
  \end{subfigure}%
  \begin{subfigure}[b]{0.5\linewidth}
    \centering\includegraphics[width=\linewidth, page=10]{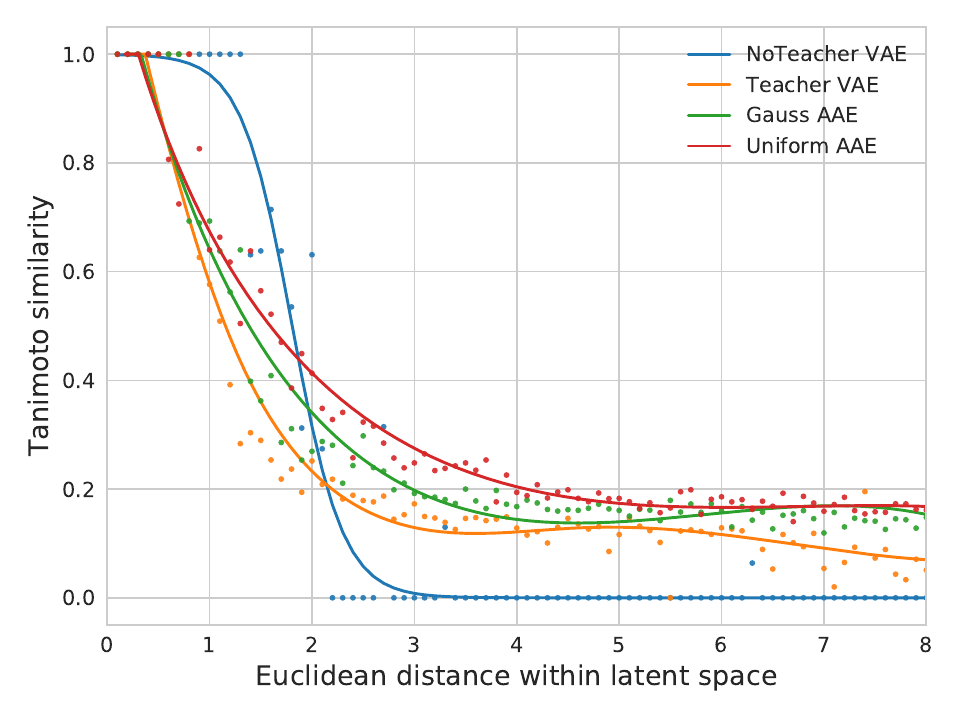}
    \caption{\label{fig:fig8b}}
  \end{subfigure}
  \caption{Results without Celecoxib in trainings set. \subref{fig:fig8a} Chemical similarity (Tanimoto, ECFP6) of generated structures to Celecoxib in relation to the distance in the latent space. \subref{fig:fig8b} Fraction of valid SMILES generated during the reconstruction.}
  \label{fig:fig8}
\end{figure*}
Celecoxib and many of its close analogues are found in the ChEMBL training set used to derive the AE models. An interesting question is whether we can maintain the smooth latent space around Celecoxib when these structures are excluded from the training set? To investigate this question, all analogues with a feature class fingerprint of diameter 4 (FCFP4) \cite{rogers_extended-connectivity_2010} Tanimoto similarity to Celecoxib larger than 0.5 (1788 molecules in total) are removed from the training set and new models are trained. The same computational process is repeated and the results are shown in Figure~\ref{fig:fig8}. 
Again Celecoxib and close analogues can be successfully reconstructed for all models at very close distance (less than 1) to Celecoxib in the latent space. In NoTeacher VAE and the Teacher VAE models, when the distance to Celecoxib is larger than 4, the fraction of valid SMILES is significantly lower compared to the Gauss and Uniform AAE models. Especially for the Uniform AAE model (shown in Figure~\ref{fig:fig8b}) at a Euclidean distance of 4, more than 30\% of the generated structures are valid SMILES. At distance of 6, the percentage of valid SMILES for the Teacher VAE is reduced to 5\% while the Uniform AAE reconstructs ~20\% of valid SMILES. This highlights that the Uniform AAE generates the smoothest latent chemical space representation. The high fraction of valid SMILES is useful for searching structures in latent space since non-decodable points in latent space make the optimization very difficult.
\setcounter{figure}{9}
\begin{figure*}[b]
    \centering
    \includegraphics[width=\linewidth, page=12]{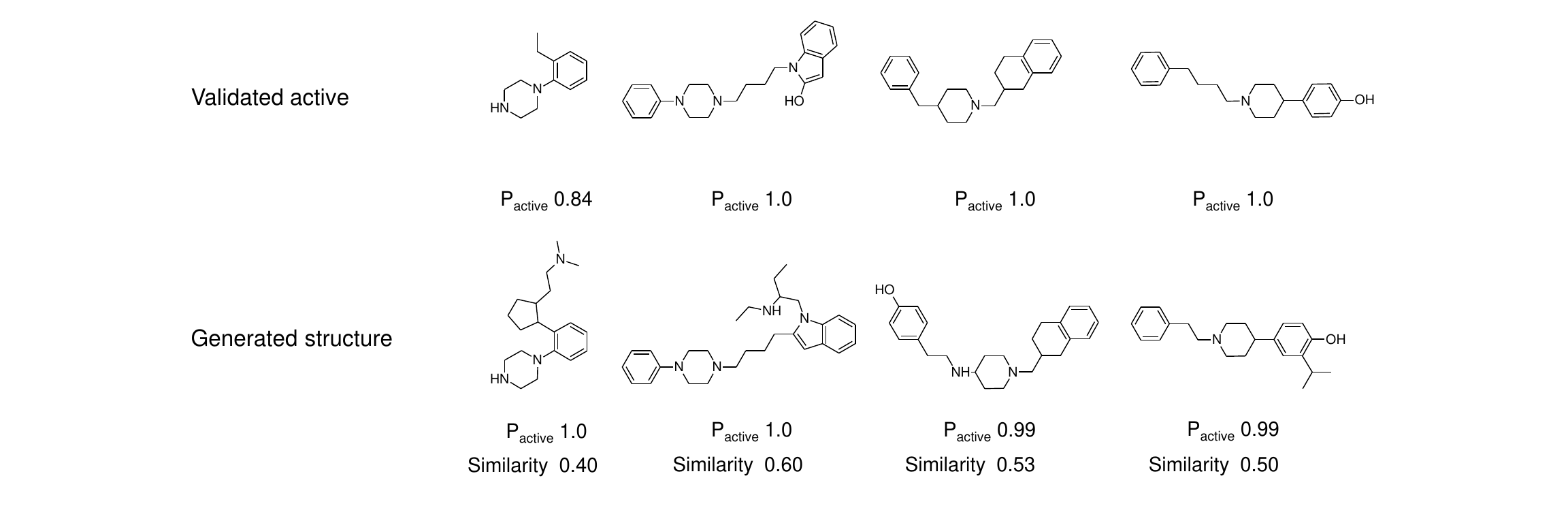}
    \caption{Generated structures from BO compared to the nearest neighbour from the set of validated actives. The validated actives were not present in the training set of the autoencoder. The Tanimoto similarity is calculated using the ECFP6 fingerprint.}
    \label{fig:fig10}
\end{figure*}
\subsection{Target-activity guided structure generation}
Given the high fraction of valid SMILES and smooth latent space for the Uniform AAE even in the case where Celecoxib is excluded from the training set, an interesting question is if we can search for novel compounds that are predicted to be active against a specific biological target. This task can be understood as an inverse-QSAR problem where one attempts to identify new compound predicted to be active by a QSAR model. We choose the DRD2 as the target using the same SVM model as above. The model is based on the ECFP6 fingerprint and the output of the SVM classifier is the probability of activity. The score for the Bayesian optimization is defined as follows:
\begin{equation*}
S(z)=
\left \{
  \begin{tabular}{p{0.73\linewidth}}
  average $P_{active}$ for all active compounds 
\\\vskip1pt
  average $P_{active}$ for all inactives if there are no active compounds
  \end{tabular}
\right .
\end{equation*}
Here we classify each compound with $P_{active} > 0.5$ as active.
During the optimization the model tends to generate structures with large macrocycles with ring sizes larger than eight atoms. This is consistent with the findings reported by G\'ómez-Bombarelli \textit{et al.} \cite{gomez-bombarelli_automatic_2016} Such large macrocycles generally have low synthetic feasibility. Thus, we include an explicit filter to remove all generated structures with ring size larger than eight.
\setcounter{figure}{8}
\begin{figure}[t]
    \centering
    \includegraphics[width=\linewidth, page=11]{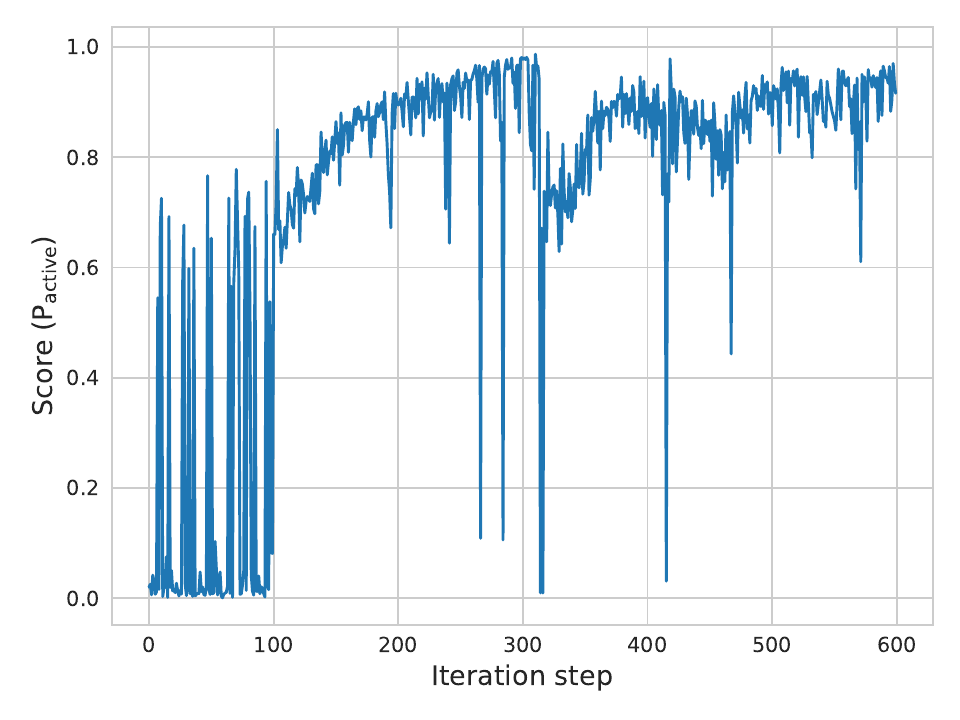}
    \caption{Searching for DRD2 active compounds using the Uniform AAE. The first 100 iterations are randomly sampled points while the next 500 iterations are determined by Bayesian optimization.\\{}\\}
    \label{fig:fig9}
\end{figure}

The BO method is used as the search engine to identify solutions with maximal scores. The BO search begins at random starting points and is repeated 80 times to collect multiple latent points with $P_{active}$ larger than 0.5. Similar results are obtained in each run. The result of a representative DRD2 search is shown in Figure~\ref{fig:fig9}. The BO algorithm constantly finds structures with high $P_{active}$ values. Low score values correspond to points at which no valid SMILES are generated or only structures with low activity scores.

In total, 369591 compounds are sampled from BO solutions with an average $P_{active}$ larger than 0.95 and 11.5\% of them have an ECFP6 Tanimoto similarity to the nearest validated active of larger than 0.35. Figure~\ref{fig:fig10} shows some exemplary compounds with a high $P_{active}$ value compared to nearest neighbours of actives. The generated compounds are predicted to be highly active ($Pactive > 0.99$) and share mostly the same chemical scaffold with the validated actives. However, the Uniform AAE model does not fully reproduce known actives. 
\setcounter{figure}{10}
\begin{figure}[t]
    \centering
    \includegraphics[width=\linewidth, page=13]{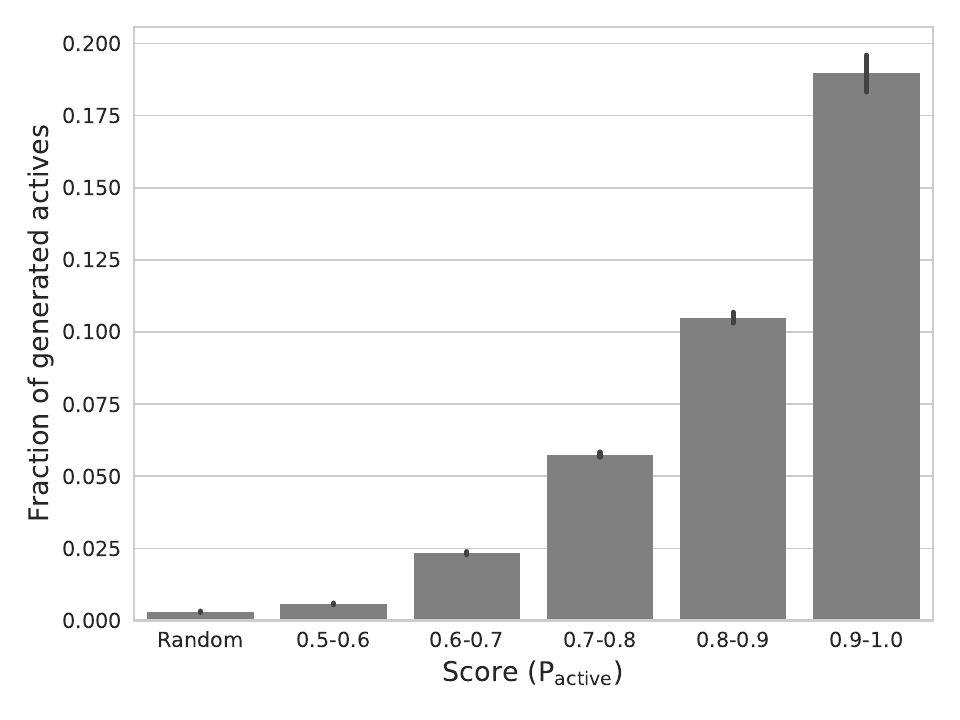}
    \caption{The relationship between the fraction of generated active compounds at specific latent points and the BO score. The fraction of generated actives is the number of actives divided by all 500 reconstruction attempts. The set “Random” corresponds to the randomly selected latent points.}
    \label{fig:fig11}
\end{figure}
The relationship between the probability of finding active compounds ($Pactive >0.5$) and score values at the latent point is shown in Figure~\ref{fig:fig11}. At latent points with higher score, there is a higher probability of finding compounds predicted to be active by the SVM model. For example, sampling at random points only yields a probability of 0.1\% to find active compounds while sampling at points with a score between 0.9 and 1.0 has a 19\% probability. This indicates that the BO algorithm can efficiently search through latent space generated by the Uniform AAE model to identify novel active compounds guided by a QSAR model.

\section{Conclusion}

In our current study, we introduce generative adversarial autoencoder NN and applied them to inverse QSAR to generate novel chemical structures. To our knowledge, adversarial autoencoder has neither been applied to structure generation nor inverse QSAR. Unlike other inverse QSAR methods that rely on back-mapping of descriptors to chemical structures, our method utilizes the latent space based generative model to construct novel structures under the guidance of a QSAR model. 
Four different AE architectures are explored for structure generation. The results indicate that sequence generation performance for novel compounds relies not only on the decoder architecture but also on the distribution of latent vectors of the encoder. The fraction of valid structures is significantly improved for architectures using teachers forcing. The autoencoder model (Uniform AAE) imposing a uniform distribution onto the latent vectors yields the largest fraction of valid structures. AE generated latent space preserves the similarity principle locally in latent space by investigating the similarity to a query structure such as Celecoxib. Furthermore, BO is applied to search for structures predicted to be active against DRD2 by a QSAR model. Our results shows that novel structures predicted to be active are identified by the BO search and this indicates that AE is a useful approach for tackling inverse QSAR problems.

\section*{Acknowledgements}
The authors thank Thierry Kogej, Ernst Ahlberg-Helgee and Christian Tyrchan for useful discussions and general technical assistance.

\section*{Competing interests}
The authors declare that they have no competing interests.

\section*{Funding}
Thomas Blaschke has received funding from the European Union’s Horizon 2020 research and innovation program under the Marie Sklodowska-Curie grant agreement No 676434, "Big Data in Chemistry" ("BIGCHEM", \url{http://bigchem.eu}). Marcus Olivecrona, Hongming Chen, and Ola Engkvist are employed by AstraZeneca. J\"urgen Bajorath is employed by University of Bonn. The article reflects only the authors’ view and neither the European Commission nor the Research Executive Agency (REA) are responsible for any use that may be made of the information it contains.

\bibliographystyle{naturemag}
\bibliography{biblio}

\end{document}